\documentclass{SPAICE}

\def\authorEmail{andrej.orsula@uni.lu}
\def\AuthorShort{Orsula et al.}

\author[]{Andrej Orsula\thanks{Corresponding author. E-Mail: \authorEmail}}
\author[]{Miguel Olivares-Mendez}
\author[]{Carol Martinez}
\affil[]{University of Luxembourg}

\title{Reward-Free Continual Adaptation for Resilient Space Robots}

\hypersetup{
    pdfinfo ={
        Title={Reward-Free Continual Adaptation for Resilient Space Robots},
        Subject={Third Conference on AI in and for Space (SPAICE 2026)},
        Author={Andrej Orsula, Miguel Olivares-Mendez, Carol Martinez},
        Creator={LaTeX},
    }
}

\begin{document}

\maketitle

\begin{abstract}%
    Space robots operate in extreme environments where hardware degradation can critically compromise traditional control strategies. While continual reinforcement learning offers a promising mechanism for online adaptation, it inherently requires access to a reward signal during deployment. However, precise reward computation in space is often infeasible due to the lack of external tracking systems and the overall complexity of the environment. To address the challenge of unobservable rewards, we introduce a reward-free continual learning framework that leverages latent-state world models. By pre-training a model-based agent across diverse simulations, the world model learns a robust predictor of the reward structure within its latent space. Upon deployment to an environment with severe hardware degradation, we freeze the observation encoder and reward predictor to update only the transition dynamics of the world model through unsupervised rollouts. By training the policy entirely on imagined trajectories generated by this updated world model, the agent adapts to altered dynamics without receiving new rewards. We demonstrate our approach across simulated planetary traversal, orbital navigation, and precision assembly tasks subjected to severe morphological failures.
    \textit{The source code is available at~\href{https://github.com/AndrejOrsula/space_robotics_bench}{https://github.com/AndrejOrsula/space\_robotics\_bench}.}
\end{abstract}

\vspace{-0.25em}
\section{Introduction}\label{sec:introduction}

Robots are positioned to deliver transformative changes for the future of space exploration and utilization. Ambitious missions envision fleets of rovers exploring unstructured planetary surfaces and mobile robotic manipulators constructing orbital megastructures~\cite{doyle2021recent}. Yet, the success of these long-duration endeavors depends heavily on developing robotic systems that can operate reliably and adapt to extreme conditions with minimal human supervision. Data-driven approaches, particularly reinforcement learning (RL), offer a powerful paradigm for acquiring such adaptive behaviors~\cite{sutton2018reinforcement}. However, a fundamental obstacle to this long-term autonomy is hardware degradation, such as the severe wheel damage observed on the Mars Curiosity rover~\cite{rankin2022assessing}. Phenomena such as thruster failures, actuator malfunctions, or kinematic misalignments can introduce significant morphological discrepancies that limit the efficacy of pre-trained control policies. When an agent encounters such severe out-of-distribution changes, zero-shot transfer would likely fail and retraining from scratch is prohibitively expensive in terms of time and communication bandwidth.

Continual RL provides a compelling alternative by allowing an agent never to stop learning and adapting its policy during deployment~\cite{abel2023definition}. Yet, the primary barrier to applying continual RL in this domain is its reliance on accurate reward signals~\cite{silver2021reward}. In simulation, computing an arbitrarily complex reward function is trivial via access to the full and perfect state. In contrast, physical robots in space cannot rely on such privileged information due to the absence of external tracking systems, limited sensing capabilities, and the unpredictable nature of the environment. Even with elaborate motion-capture systems within terrestrial analogue facilities, estimating rewards might be practically infeasible for many critical tasks. For instance, the rewards and penalties for autonomous regolith excavation can be computed in simulation by directly analyzing the positions and velocities across millions of discrete particles~\cite{orsula2025learningtool}. However, accurately estimating the volume of excavated regolith and ejected dust in a real-world lunar environment would be rather challenging and unreliable~\cite{cloud2021towards}. This fundamental challenge of unobservable rewards has been a critical bottleneck to deploying adaptive learning systems across most real-world domains, particularly in space robotics, where the stakes of failure are high, and opportunities for human intervention are limited.

\begin{figure}[t]
    \vspace{0.35em}
    \centering
    \includegraphics[width=1.0\linewidth]{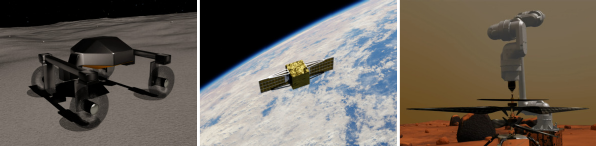}
    \caption{Our continual learning framework enables reward-free adaptation to severe changes in transition dynamics by leveraging the latent reward landscape encoded within a pre-trained world model. We demonstrate this approach across three distinct space robotics domains, where the agent recovers from severe failures.}
    \label{fig:teaser}
    \vspace{-1.0em}
\end{figure}

Recent advancements in model-based RL, particularly latent-state world models like DreamerV3~\cite{hafner2025mastering}, have demonstrated significant potential for sample-efficient robot learning. Concurrently, studies have shown that world model dynamics can be pre-trained using offline datasets and fine-tuned on real-world data~\cite{feng2023finetuning}. Despite these developments, most existing approaches continue to rely on either manual or meticulously engineered sparse reward signals while primarily addressing domain shifts. Our focus lies in addressing the challenge of unobservable rewards during continual adaptation to severe changes in the underlying transition dynamics. By leveraging the latent representation learned during pre-training in diverse simulations, we hypothesize that world models inherently encode a robust reward landscape within their latent space. Subsequent online adaptation can then be achieved by updating only the transition dynamics of the world model through unsupervised rollouts while freezing the observation encoder and reward predictor. In parallel, the active policy can be updated entirely on synthetic trajectories generated by the updated world model, which allows the agent to adapt to changing dynamics without receiving new rewards.

This work introduces a framework for reward-free continual adaptation using world models. We demonstrate the efficacy of our approach across three distinct space robotics domains: planetary traversal, orbital navigation, and precision assembly. In each case, we simulate severe morphological failures in the form of actuator malfunctions or kinematic misalignments to introduce significant discrepancies between the pre-trained and degraded dynamics. Under time and interaction constraints, we evaluate the performance of our agent against a zero-shot baseline, an agent with privileged reward information, and one retrained from scratch. This work represents a step towards enabling truly resilient autonomy for extreme environments, where robots can adapt to unforeseen challenges without human intervention.

\vspace{-0.25em}
\section{Methodology}\label{sec:methodology}

Our framework for reward-free continual adaptation builds upon the DreamerV3 architecture~\cite{hafner2025mastering}, which leverages a latent-state world model to compress high-dimensional observations into compact representations. During the pre-training phase, we follow the standard procedure of jointly optimizing the world model on observed trajectories while simultaneously training an actor-critic policy entirely within the synthesized rollouts of the world model.

\begin{figure}[ht]
    \centering
    \includegraphics[width=0.925\columnwidth]{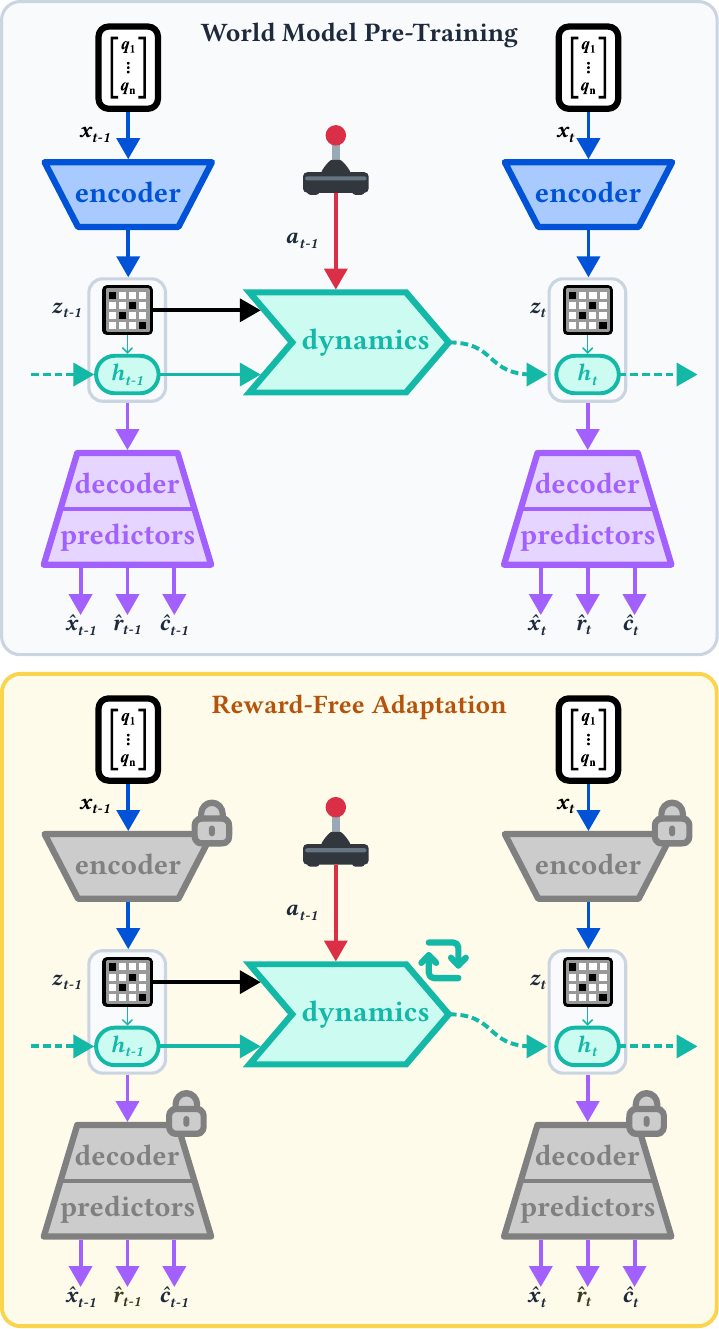}
    \caption{The world model is pre-trained in simulation. Once deployed, only the transition dynamics are updated via rollouts.}
    \label{fig:latent_concept}
    \vspace{-1.0em}
\end{figure}

The core scientific exploration of our methodology lies in the online adaptation phase. As illustrated in \cref{fig:latent_concept}, we strictly update only the transition dynamics of the world model while freezing the encoder and other heads, including the reward predictor. With this minimal change, the agent can leverage the latent reward landscape learned during pre-training to guide its adaptation to the new environment dynamics. The workflow is therefore divided into two distinct phases: world model pre-training and reward-free adaptation.

\subsection{World Model Pre-Training}

During simulation pre-training, the world model learns to encode high-dimensional observations~$x_t$ into compact stochastic latent states~$z_t$. The Recurrent State-Space Model (RSSM) of DreamerV3 is defined by the following core components~\cite{hafner2025mastering}:
{\setlength{\jot}{1pt}
\begin{align}
    \text{Sequence Model:} \quad       & h_t = f_\phi(h_{t-1}, z_{t-1}, a_{t-1}) \label{eq:seq}         \\
    \text{Forward Dynamics:} \quad     & \hat{z}_t \sim p_\phi(\hat{z}_t \mid h_t) \label{eq:dyn}       \\
    \text{Encoder:} \quad              & z_t \sim q_\phi(z_t \mid h_t, x_t) \label{eq:enc}              \\
    \text{Decoder:} \quad              & \hat{x}_t \sim p_\phi(\hat{x}_t \mid h_t, z_t) \label{eq:dec}  \\
    \text{Reward Predictor:} \quad     & \hat{r}_t \sim p_\phi(\hat{r}_t \mid h_t, z_t) \label{eq:rew}  \\
    \text{Continuity Predictor:} \quad & \hat{c}_t \sim p_\phi(\hat{c}_t \mid h_t, z_t) \label{eq:cont}
\end{align}}
The model is trained jointly to reconstruct observations~$\hat{x}_t$, predict rewards~$\hat{r}_t$ and episode continuity~$\hat{c}_t$, while minimizing the Kullback-Leibler (KL) divergence between the prior dynamics (\cref{eq:dyn}) and the posterior representations (\cref{eq:enc}). Concurrently, an actor-critic policy~$\pi(a_t \mid h_t, \hat{z}_t)$ is trained entirely within latent trajectories generated by the RSSM. The critic is optimized using the predicted rewards from the reward head, which encourages the world model to learn a latent representation that captures the underlying reward structure of the task.

To ensure that the learned latent representations and reward mappings are robust across the vast distribution of possible states, we employ domain randomization during this pre-training~\cite{tobin2017domain}. This encourages the encoder to learn a highly generalizable mapping into the latent space and ensures the reward head accurately reflects the true objective function that remains invariant to perturbations.

\subsection{Reward-Free Adaptation}

Upon deployment, it is anticipated that any robot will eventually encounter novel environmental conditions and hardware degradation that significantly alter the underlying transition dynamics. As the actor-critic policy was trained entirely within imagination, its behavior inherently captures the transition dynamics encoded in the original world model. Therefore, any significant divergence in the true dynamics of the new environment would lead to a catastrophic failure of the zero-shot policy as the sampled actions would no longer yield the expected outcomes.

Online adaptation is therefore critical to recover performance. We address the key challenge of unobservable rewards by leveraging the fact that the reward predictor (\cref{eq:rew}) was pre-trained to capture the underlying reward structure within the latent space of the world model. By freezing this head during adaptation, we expect the actor-critic to continue receiving meaningful reward signals based on its updated latent representations, even as the transition dynamics change. This allows the agent to adapt its policy purely through unsupervised rollouts that update the transition dynamics while maintaining a consistent reward landscape.

In addition to freezing the reward predictor, we also freeze the encoder (\cref{eq:enc}) and decoder (\cref{eq:dec}) to preserve the integrity of the latent representations. By updating only the sequence model (\cref{eq:seq}) and the forward dynamics (\cref{eq:dyn}), we can ensure that the agent learns a corrective delta to the transition dynamics without degrading the foundational world model. We optimize the transition dynamics using the KL divergence loss between the posterior and prior distributions on the new proprioceptive transitions:
\begin{equation}
    \mathcal{L}_{dyn} \doteq \mathrm{KL}(q_\phi(z_t \mid h_t, x_t) \parallel p_\phi(\hat{z}_t \mid h_t)) \label{eq:kl_loss}
\end{equation}
To mitigate catastrophic forgetting, we reduce the learning rate of the world model by an order of magnitude ($4 \hspace{-0.75mm} \times \hspace{-0.75mm} 10^{-5} \rightarrow 4 \hspace{-0.75mm} \times \hspace{-0.75mm} 10^{-6}$). Furthermore, we inject a small Gaussian exploration noise $\mathcal{N}(0.0, 0.02)$ into the normalized output actions during online rollout collection to ensure the RSSM experiences novel transitions under the new physical constraints. The actor-critic is then repeatedly re-trained in imagination using the newly updated dynamics with a training ratio of $2048$ policy updates per environment step.

For fixed-horizon tasks, the continuity predictor (\cref{eq:cont}) can either remain frozen or be updated to reflect new episode termination conditions. We freeze it since termination conditions remain unchanged across our pre-training and deployment environments. However, if degradation significantly alters the episode structure, updating this predictor may be necessary.

\vspace{-0.25em}
\section{Experimental Results}
\label{sec:results}

We conduct a series of simulation-only experiments to demonstrate key aspects of our adaptation framework.

\begin{figure*}[t]
    \centering
    \includegraphics[width=\linewidth]{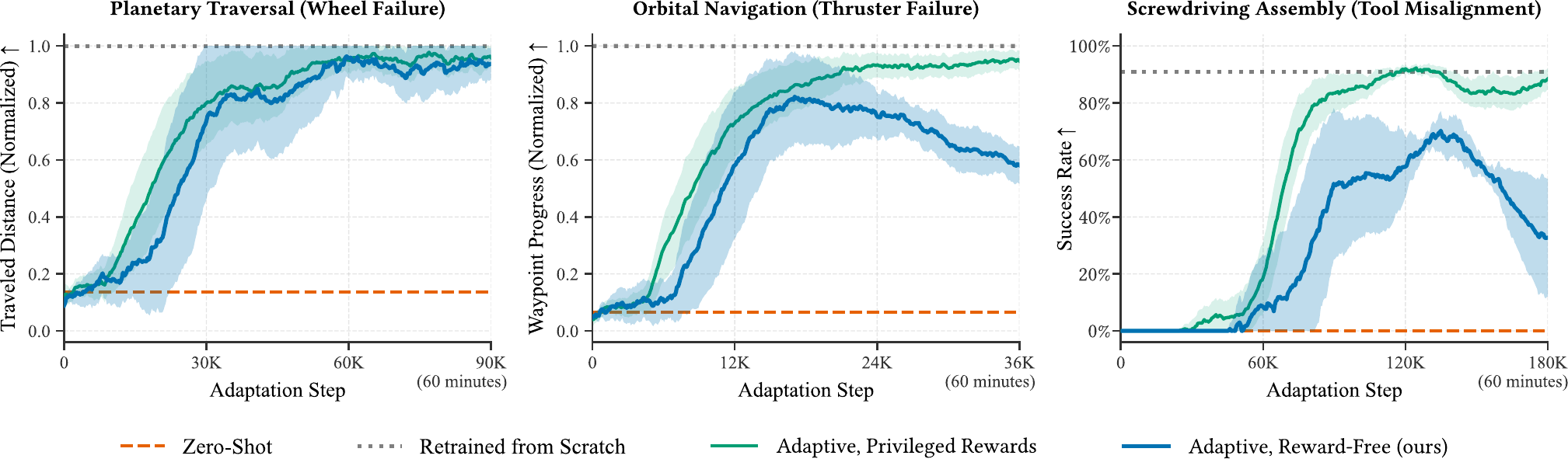}
    \caption{Performance curves for the three tasks over a 60-minute online adaptation window. The adaptive agent with access to privileged rewards reaches performance levels close to the agent retrained from scratch. Our reward-free agent also rapidly recovers performance, albeit to a lower degree and with decay. The zero-shot agent fails catastrophically due to the unmodeled dynamics shift.}
    \label{fig:learning_curves}
    \vspace{-1.0em}
\end{figure*}

\subsection{Task Formulation and Failure Modes}

As conceptualized in \cref{fig:teaser}, we designed three distinct tasks across planetary and orbital domains, each paired with an unmodeled morphological failure that induces a significant shift in the underlying transition dynamics. All tasks are implemented on top of NVIDIA Isaac Lab~\cite{mittal2025isaaclab} within the scope of the Space Robotics Bench~\cite{orsula2025space}.

\vspace{-0.85em}
\paragraph{Planetary Traversal (Wheel Failure) [25 Hz]}%
Traversal across complex terrain with an experimental rover inspired by NASA JPL's ERNEST prototype~\cite{nguyen2024trajectory}. Our 12-actuator rover features a double-sided active gimbal suspension system for maintaining ground contact and stability across extreme terrain. The rover is tasked with traversing a procedurally generated obstacle course while maintaining a target velocity vector. A failure mode is simulated by locking the steering and drive joint of the front-right wheel, which introduces significant asymmetric drag.

\vspace{-0.85em}
\paragraph{Orbital Navigation (Thruster Failure) [10 Hz]}%
Dynamic waypoint navigation in $\mathrm{SE}(3)$ under microgravity using a spacecraft with 12 independent continuous thrusters. The agent must track a moving target pose while compensating for the dynamics of spaceflight. We simulate the complete failure of three co-located off-axis thrusters, which results in a shift of the transition dynamics.

\vspace{-0.85em}
\paragraph{Screwdriving Assembly (Tool Misalignment) [50 Hz]}%
Assembly of a pre-aligned bolt into a matching nut using a 7-DOF robotic manipulator that is equipped with an electric screwdriver end-effector. The task requires precise alignment and insertion under tight tolerances. For the failure mode, we introduce a 15\textdegree\ axial bend in the mounting flange of the screwdriver, which causes the tool tip to deviate from the kinematic expectations.

\subsection{Experimental Protocol}\label{subsec:protocol}

Although the evaluation is performed solely inside a simulation, we enforce a strict separation between the data-abundant simulation phase and the resource-constrained adaptation phase. During pre-training, the model-based agent is trained for $20$ million environment steps, with a training ratio of $32$ policy updates per step across $512$ parallel environment workers while randomizing physical parameters, such as the gravity vector, inertial properties, friction coefficients, and random external disturbances. This diverse parallelization ensures the world model acquires a general latent representation of both the physical dynamics and the underlying reward landscape.

In contrast, the subsequent online adaptation phase is strictly constrained to a single simulated agent environment. This constraint is crucial for accurately mimicking the severe data-collection bottleneck of a physical robot deployed in an isolated extraterrestrial environment. Furthermore, we limit the adaptation window to exactly $60$~minutes of interaction time. Depending on the control frequency of the respective robotic platforms, this translates to $90$K steps for planetary traversal, $36$K steps for orbital navigation, and $180$K steps for screwdriving assembly.

Throughout both phases, the agent maintains access to proprioceptive observations alongside the state for relative target tracking (planetary traversal and orbital navigation) or the pose of the bolt (screwdriving assembly). Furthermore, the action space remains unchanged even after morphological degradation that renders certain actions obsolete due to actuator failure. For each task, the adaptation phase is repeated across three random seeds with identical pre-trained models to evaluate the robustness.

\vspace{-0.25em}
\subsection{Performance Evaluation}\label{subsec:evaluation}

We track normalized task-specific progress metrics and evaluate four agents: \textbf{a)} a zero-shot baseline that evaluates the pre-trained policy; \textbf{b)} an agent retrained from scratch on the new dynamics to serve as an asymptotic upper bound; \textbf{c)} an adaptive agent with access to privileged rewards; and \textbf{d)} our reward-free adaptive agent. Aside from the zero-shot baseline, our agent is the only adaptive baseline that lacks access to the true reward signal and relies entirely on the latent reward landscape encoded within the world model, making it viable for real-world applications.

As illustrated in \cref{fig:learning_curves}, the zero-shot policies fail catastrophically across all domains due to the unmodeled dynamics shift. While the retrained baseline confirms that the tasks are solvable under hardware degradation, it highlights the extreme sample inefficiency of learning from scratch. The privileged agent demonstrates rapid recovery, achieving performance nearing the retrained upper bound.

Conversely, our reward-free agent also shows promising initial recovery, demonstrating that latent world models can guide adaptation without external rewards. However, the learning profiles reveal clear limitations. In all three domains, our agent consistently underperforms compared to the privileged baseline. Notably, after an initial performance gain, the agent exhibits significant volatility and decay, particularly in the orbital and assembly tasks. These results suggest that while the latent reward landscape is sufficient for short-term policy recovery, it lacks the long-term stability of explicit rewards, as updating the transition dynamics on degraded morphologies likely causes the RSSM representation to drift away from the original latent space.

\vspace{0.5em}
\section{Discussion and Conclusion}\label{sec:conclusion}

The experimental results validate our central hypothesis that latent-state world models pre-trained across diverse simulations encode a reward landscape that can guide online adaptation without observing new rewards. By isolating the dynamics update from the reward prediction, this framework resolves a critical bottleneck in space robotics, where computing online rewards is often impossible.

However, the late-stage decay exposes a capacity limit within the RSSM. Because the reward head remains frozen, continuously updating the core transition dynamics on degraded morphologies eventually overwrites the generalized principles acquired during pre-training.

Despite its effectiveness, our framework highlights clear limitations. To prevent late-stage decay, future work will investigate localized latent-space adapters that strictly bound dynamics updates. Notably, our simulation-only study bypasses the peculiarities of the sim-to-real gap~\cite{salvato2021crossing}. Furthermore, the demanding optimization of the online adaptation phase heavily exceeds the strict power constraints of space-grade embedded compute modules~\cite{felix2024total}, necessitating future breakthroughs in efficient in-situ learning.

\addcontentsline{toc}{section}{References}
\printbibliography

\end{document}